\documentclass[10pt,english]{article}
\usepackage{subfigure}
\usepackage{graphicx}
\usepackage{float}
\usepackage[T1]{fontenc}
\usepackage[latin9]{inputenc}
\usepackage[margin=1.25in]{geometry}
\usepackage{float}
\usepackage{amsmath}
\usepackage{amsbsy}
\usepackage{setspace}
\usepackage{amssymb}
\usepackage{esint}
\usepackage{cite}
\newfloat{algorithm}{tbp}{loa}
\floatname{algorithm}{Algorithm}
\usepackage{hyperref}
\usepackage{listings}

\newlength\myindent
\makeatletter

\floatstyle{ruled}
\newfloat{algorithm}{tbp}{loa}
\floatname{algorithm}{Algorithm}

\usepackage{babel}

\begin{document}
%\begin{singlespace}

\title{TensorFlow Chaotic Prediction and Blow Up}

\author{M. Andrecut}

\date{September 14, 2023}

\maketitle
{

\centering Unlimited Analytics Inc.

\centering Calgary, Alberta, Canada

\centering mircea.andrecut@gmail.com

} 
\begin{abstract}

Predicting the dynamics of chaotic systems is one of the most challenging tasks for neural networks, and machine learning in general. 
Here we aim to predict the spatiotemporal chaotic dynamics of a high-dimensional non-linear system. 
In our attempt we use the TensorFlow library, representing the state of the art for deep neural networks 
training and prediction. While our results are encouraging, and show that the dynamics of the considered system can be predicted for short time, 
we also indirectly discovered an unexpected and undesirable behavior of the TensorFlow library. More specifically, the longer term prediction of the system's chaotic 
behavior quickly deteriorates and blows up due to the nondeterministic behavior of the TensorFlow library. Here we provide numerical evidence 
of the short time prediction ability, and of the longer term predictability blow up.


\end{abstract}

\section{Introduction}

Predicting the dynamics of complex systems exhibiting high-dimensional spatiotemporal chaos is a challenging machine learning problem with important applications in: 
physics, biology, medicine, economics, meteorology etc. 
Another problem of interest, is the inverse problem of inferring the connectivity network of such a system from input-output measurements. 
Such an example is the case of inferring the connectivity of genetic regulatory networks from the measurements of gene expression data. 
Here we explore the feasibility of these problems using a complex system corresponding to a non-linear network 
model we previously discussed in \cite{key-1}. This is a continuous model of non-linear random networks (NLRN), which exhibits a 
phase transition from ordered to chaotic dynamics as a function of the average network connectivity (in-degree). 
In the chaotic regime, these networks show strong sensitivity to initial conditions, quickly forgetting their past states, 
making them harder to predict and to infer their connectivity. In our approach we use the TensorFlow library \cite{key-2}, which is the state of the art for deep neural networks 
training and prediction. Our numerical results show that the dynamics of the considered system can be successfully predicted for short times. 
However, we also indirectly discovered that longer term predictions of the chaotic system quickly deteriorate and blow up due to an 
unexpected behavior of the TensorFlow library. Here we provide numerical evidence 
of the short time prediction ability, and of the longer term predictability blow up.

\section{The NLRN model}

The NLRN model considered here is a slight variation of the model discussed in \cite{key-1}, and it consists of $N$ randomly interconnected variables, 
with continuously valued states $x_n \in (-1,1)$, $n=0,1,...,N-1$. At time step $t$ the 
state of the network is described by the $N$ dimensional activity vector:
\begin{equation}
\mathbf{x}(t) = [x_0(t),x_1(t),...,x_{N-1}(t)]^T,
\end{equation}
which is iteratively updated at time $t+1$ using the following equation:
\begin{equation}
\mathbf{x}(t+1) = f(\mathbf{u},\mathbf{a},\mathbf{x}(t)), 
\end{equation}
where $f$ is a non-linear activation function. In this paper we assume that $f$ is the $\text{tanh}()$ function:
\begin{equation}
f_n(\mathbf{u},\mathbf{a},\mathbf{x}(t)) = \text{tanh} \left( \sum_{m=0}^{N-1} u_{nm}x_m(t) + a_n \right), \quad n = 0,1,...,N-1.
\end{equation}
Here, $\mathbf{u}$ is the network connectivity matrix and $\mathbf{a}$ is the bias vector, with randomly assigned elements $u_{nm}, a_n \in (-1,1)$. 
That is, we assume a fully connected network, which is known to exhibit spatiotemporal chaotic behavior. 
The problems we would like to solve are: (1) to predict the chaotic dynamics of the system, and (2) to infer the connectivity matrix $\mathbf{u}$ and the bias vector $\mathbf{a}$, using only 
the observed activity vector $\mathbf{x}(t)$. 

One can see that the NLRN model corresponds to a chaotic single layer feed-forward neural network with frozen weights and biases. 
It is therefore natural to consider the problem of another single layer feed-forward neural network trying to learn the connectivity matrix and the bias vector of the given chaotic network. 

Therefore, we assume a similar model with the connectivity matrix $\mathbf{w}$, and bias $\mathbf{b}$:
\begin{equation}
\tilde{x}_n(t+1) = \text{tanh} \left( \sum_{m=0}^{N-1} w_{nm}\tilde{x}_m(t) + b_n \right), \quad n = 0,1,...,N-1,
\end{equation}
and we aim to learn $\mathbf{w}$ and $\mathbf{b}$ using the observed dynamics of the given chaotic model (3).

The procedure we will use consists of iterating the chaotic system (3) for a number of $T>0$ time steps, starting from a randomly generated initial state: $x_n(0) \in (-1,1)$, $n=0,1,...,N-1$. 
We collect the pairs of input-output state vectors $(\mathbf{x}(t), \mathbf{x}(t+1))$, and we train the neural network defined by (4) in order to find $\mathbf{w}$ and $\mathbf{b}$, and then  
to infer $\mathbf{u}\leftarrow \mathbf{w}$ and $\mathbf{a}\leftarrow \mathbf{b}$, and to make predictions. 
Our expectations are that after training we will obtain a good approximation of the unknown network, $\mathbf{w} \approx \mathbf{u}$ and $\mathbf{b} \approx \mathbf{a}$, 
and a good prediction of the future dynamics, $\mathbf{\tilde{x}}(t) \approx \mathbf{x}(t)$, for $t>T$. 

\begin{figure}[!ht]
\centering \includegraphics[width=10cm]{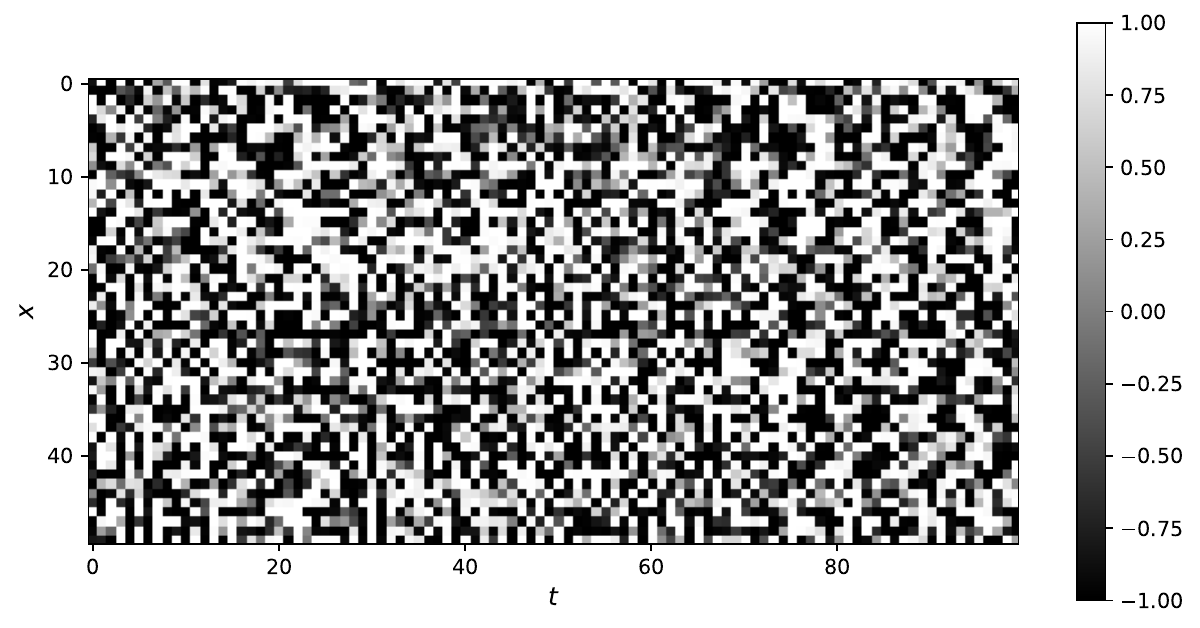}
\centering \includegraphics[width=10cm]{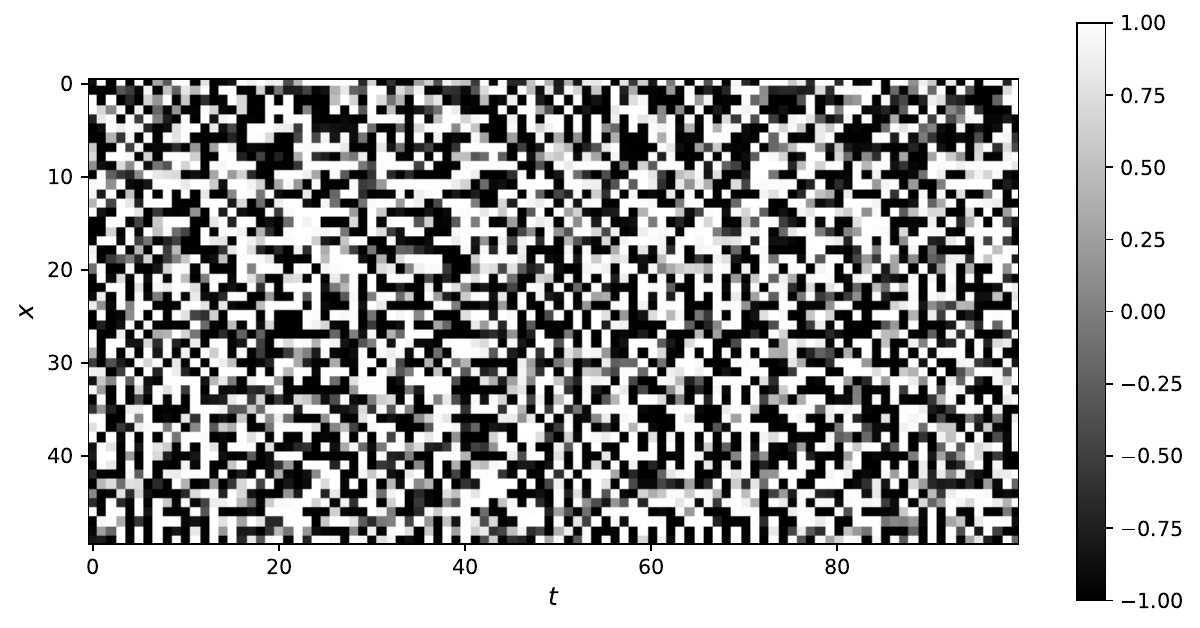}
\centering \includegraphics[width=10cm]{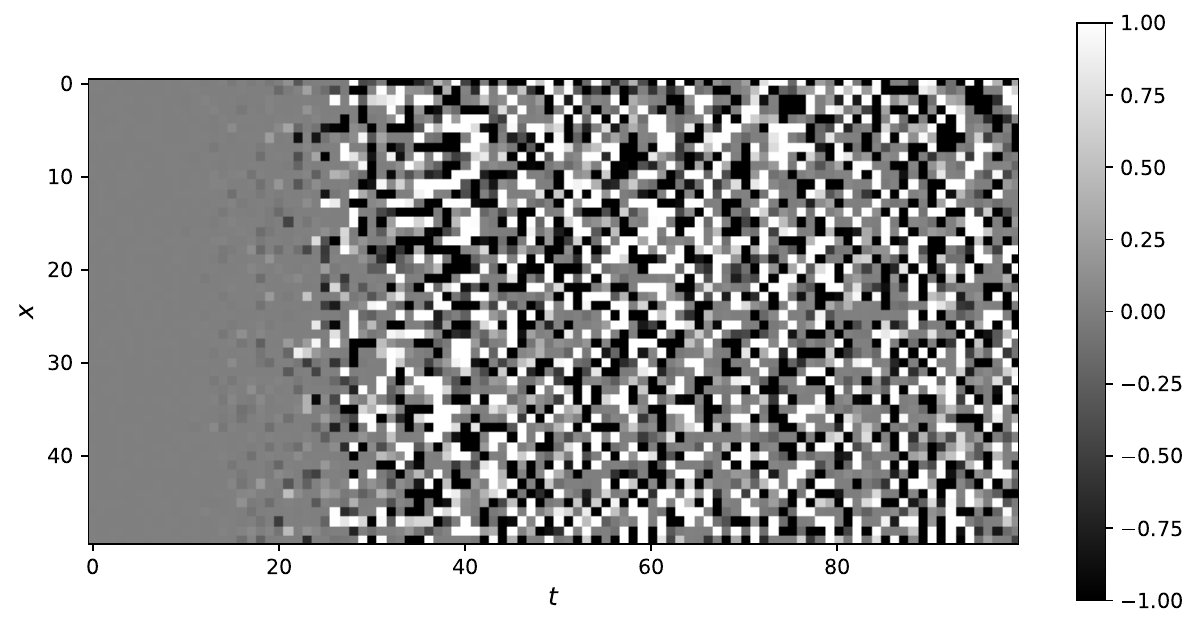}
\caption{Prediction results for $N=50$ and $\Delta T = 100$ (see the text for details).}
\end{figure}

\section{TensorFlow inference and prediction blow up}

The TensorFlow model used here is probably the simplest one, since it consists of a single layer neural network with a $\text{tanh}()$ activation function \cite{key-3},\cite{key-4}:
\begin{small}
\begin{verbatim}
model = keras.Sequential()
model.add(layers.Input(shape=(N,)))
model.add(layers.Dense(N, activation="tanh"))
model.compile(loss=keras.losses.MeanSquaredError(),optimizer=keras.optimizers.Adam(0.001))
model.fit(x,y, epochs=ne, batch_size=nb)
\end{verbatim}
\end{small}
Here, 'ne' is the number of epochs used for training, and 'nb' is the number of samples included in each training batch. 
We also used the 'MeanSquaredError()' loss function with the Adam optimizer, and a learning rate of 0.001. 

In a first example, we iterated an instance of the chaotic network with $N=50$ nodes for $T=1000$ steps, collecting the input-output pairs, then we trained the TensorFlow network for nepochs=1000 and nbatch=100. 
In Figure 1 we give the results of such a prediction exercise for a time $t \in [T,T+\Delta T]$, where $\Delta T = 100$. On top we have the real 'future' dynamics of the chaotic system (simulated using the standard Numpy library), 
in the middle we have the predicted dynamics of the TensorFlow learned neural network, and at the bottom we have the difference of the two systems. One can see that the prediction is quite good up to 
about $t^* \simeq 15$ steps, when it abruptly starts to deteriorate. 
We should also note that in this particular case the relative errors for the weights and bias are quite small: 
\begin{equation}
\varepsilon_{w} = 100\frac{\Vert \mathbf{w} - \mathbf{u} \Vert}{\Vert \mathbf{u} \Vert} = 0.0722 \%, \quad \varepsilon_{b} = 100\frac{\Vert \mathbf{b} - \mathbf{a} \Vert}{\mathbf{\Vert a \Vert}} = 0.0587 \%, 
\end{equation}
and therefore the targeted network is inferred very well. However, the prediction error deteriorates unexpectedly after just a few steps. 
This observation triggered the following question: If the inference errors are so small then why the prediction deteriorates after such a small number of steps? 
In fact we observed this unexpected phenomenon even when the inference errors were zero, which obviously should not happen, since in this case the chaotic and the 
learned systems should be identical. 

Once a model is trained, TensorFlow uses the 'model.predict()' method to predict the output for a given input:
\begin{small}
\begin{verbatim}
x[0] = x0
for t in range(T):
    x[t+1] = model.predict(x[t])
\end{verbatim}
\end{small}
here $x0$ is the initial condition. This prediction method should be deterministic. That is, for a given input $x(t)$ the output $x(t+1)$ should always be the same. 
What makes it unpredictable after just a few iteration steps?

\begin{figure}[!ht]
\centering \includegraphics[width=10cm]{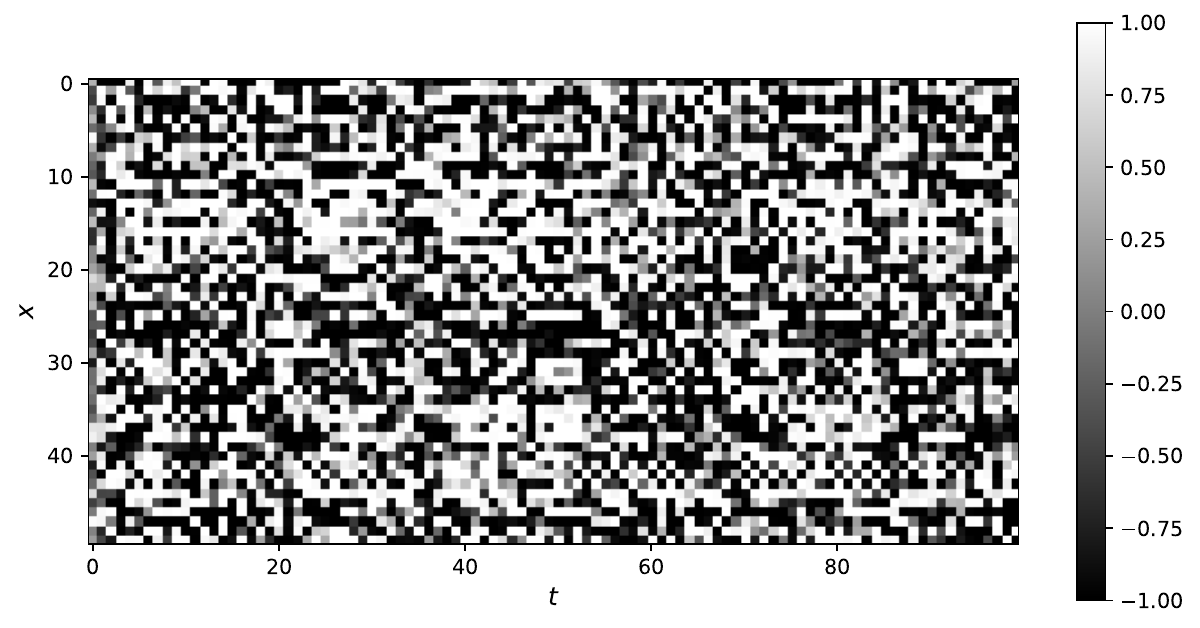}
\centering \includegraphics[width=10cm]{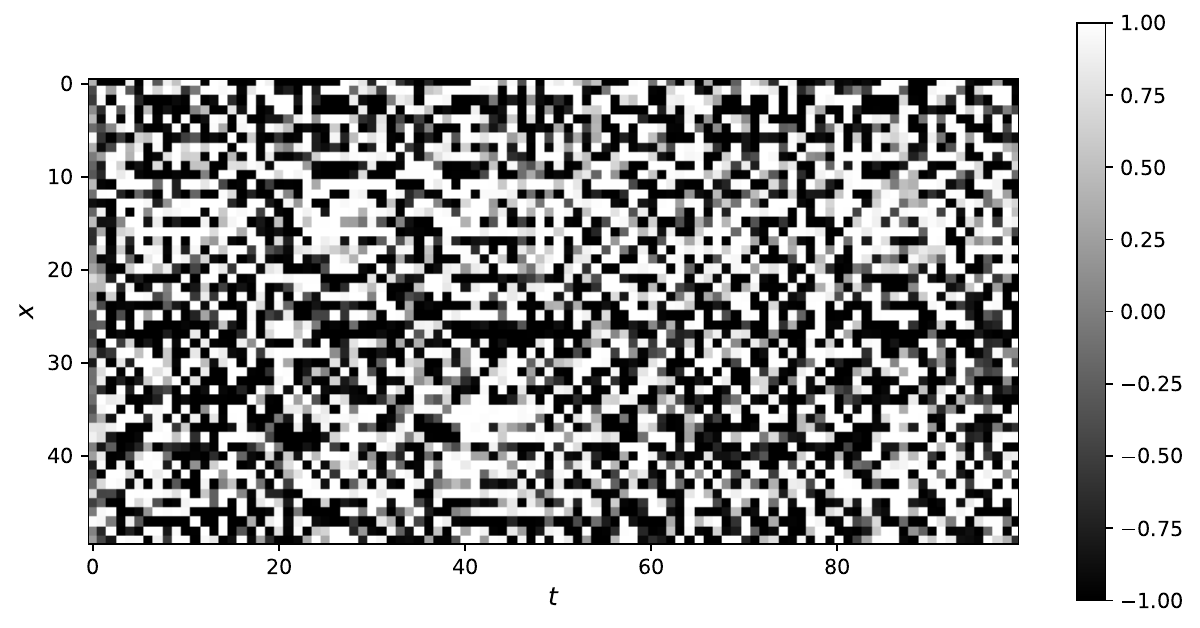}
\centering \includegraphics[width=10cm]{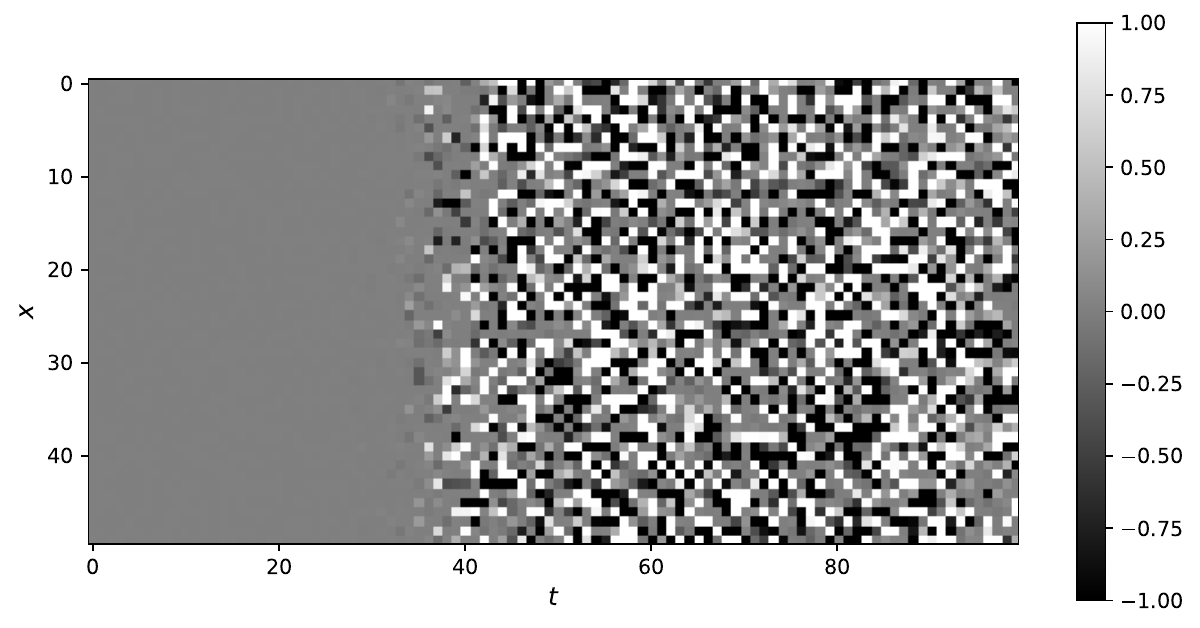}
\caption{Prediction results for $N=50$ and $\Delta T = 100$: Numpy system (top); Tensorflow system (middle); Difference system (bottom).}
\end{figure}

These observations prompted us to investigate further, and to simplify the investigation we assumed the same randomly generated weights and biases for both systems, that 
is we removed the learning step from the process and we simply assumed that $\mathbf{w}=\mathbf{u}$, and $\mathbf{b}=\mathbf{a}$. We iterated both systems starting from the same randomly generated 
initial condition (the code is given in Appendix 1). The first system was iterated using the Numpy library, while the second system was iterated using the TensorFlow library. 
The results are shown in Figure 2 for $N=50$, $T=100$. The top figure shows the system iterated using the Numpy library, the middle figure shows the system iterated with the TensorFlow library, 
while the bottom figure shows the dynamics difference of the two systems. 
One can clearly see that after a few iterations the dynamics of these systems diverges completely, even though theoretically speaking they should be identical. 
In this case the TensorFlow system blow up takes a bit longer to set up, since in the previous example in Figure 1, the systems were not quite identical, as a result of the learning process being present. 
However, this behavior of TensorFlow is quite unexpected and disturbing, since many problems require a prediction as accurate as possible. 

We repeated the experiment with a double number of nodes, $N=100$, and surprisingly the TensorFlow system iterations blow up even earlier, and more abruptly, 
suggesting that this undesirable and unexpected phenomenon also depends on the 
number $N$ of nodes in the neural network. In Figure 3 we show how this TensorFlow chaotic blow up transition depends on the number of nodes in the neural network for: $N=50$, $N=100$, $N=1000$ and $N=10000$. 
Here we have calculated the error between the Numpy and TensorFlow computed states at each iteration step, and we averaged over 100 trials to have a better statistical estimation (the code is listed in Appendix 2). 

\begin{figure}[!ht]1
\centering \includegraphics[width=9.5cm]{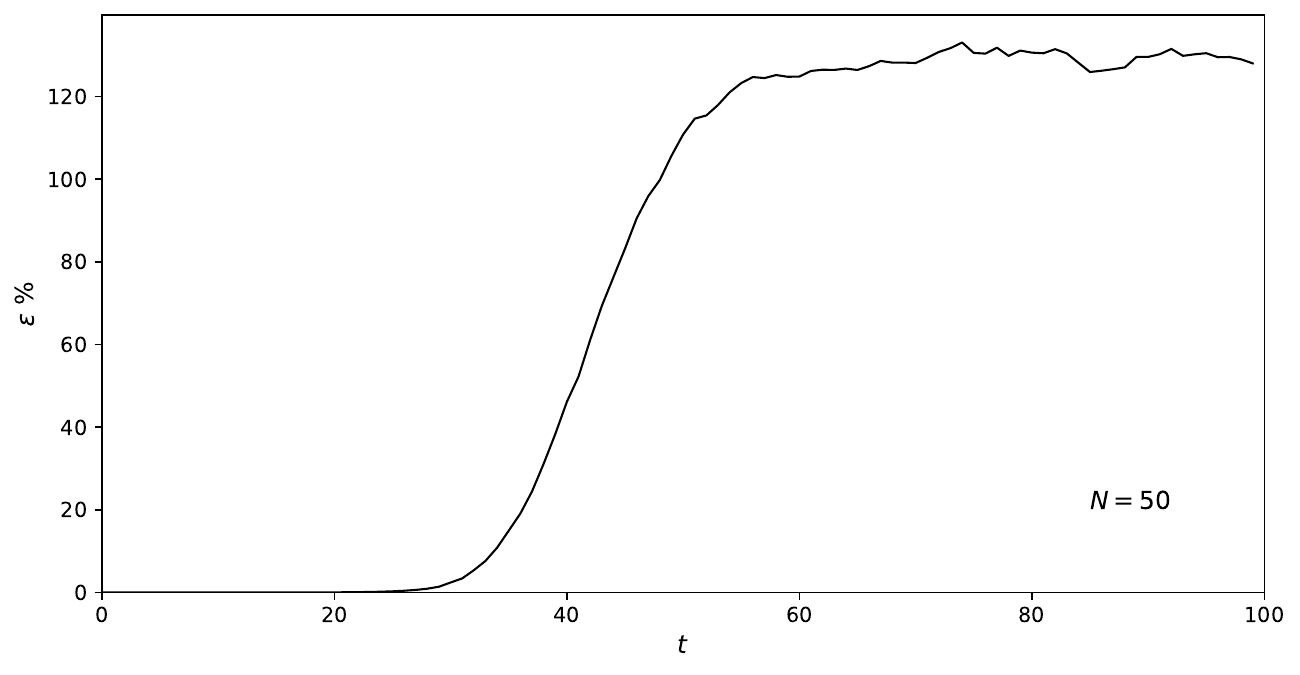}
\centering \includegraphics[width=9.5cm]{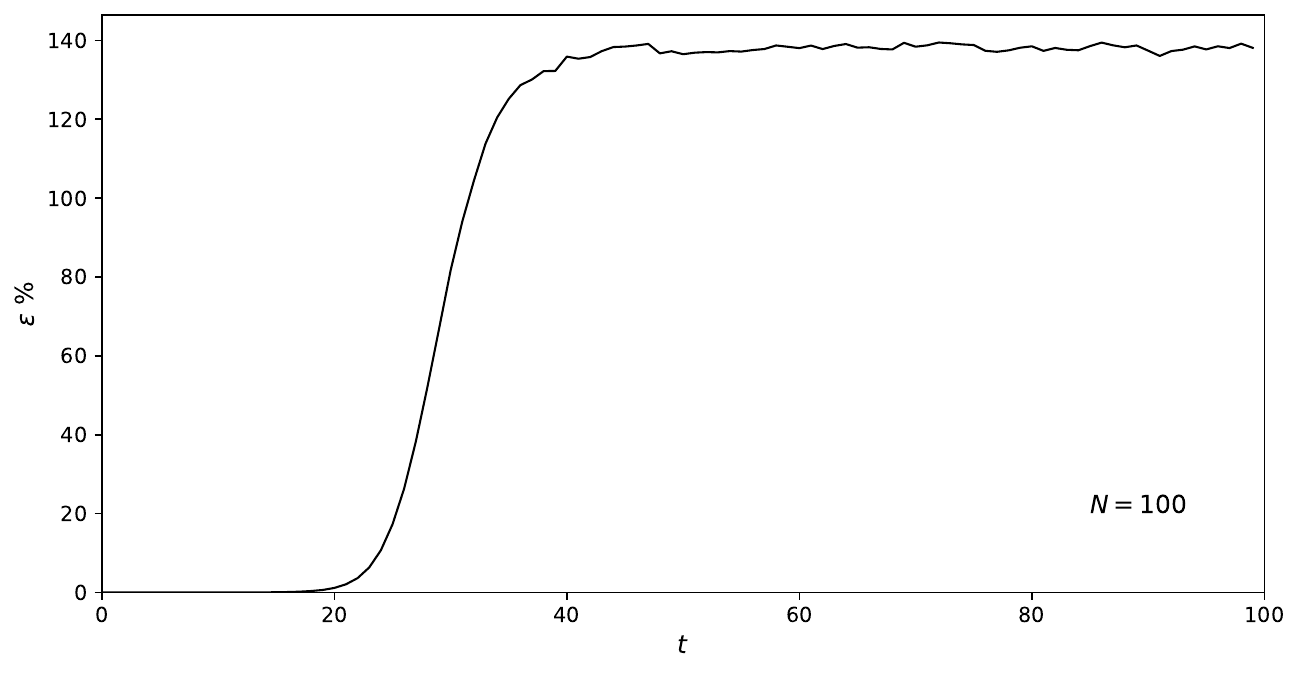}
\centering \includegraphics[width=9.5cm]{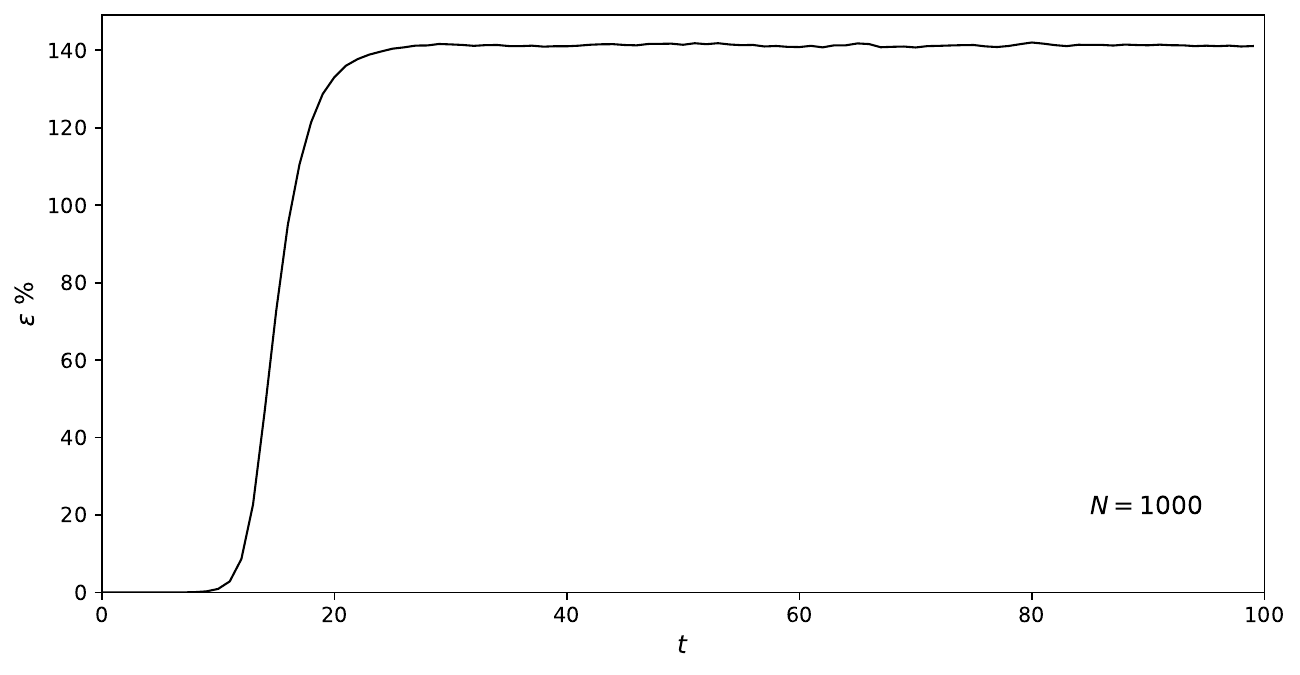}
\centering \includegraphics[width=9.5cm]{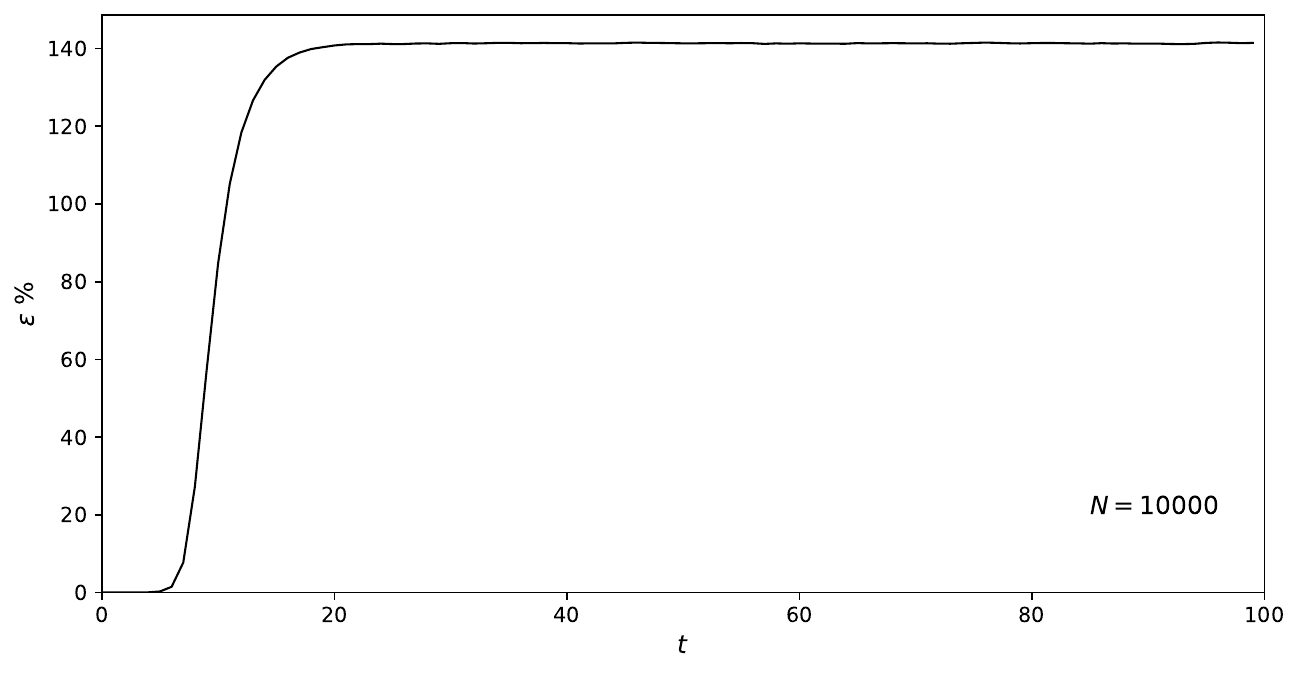}
\caption{Tensorflow prediction blow up for: $N=50$, $N=100$, $N=1000$ and $N=10000$ nodes.}
\end{figure}

One can see that by increasing the number of nodes $N$ the Tensorflow system iteration blows up earlier, and more abruptly, suggesting that larger systems cannot be reliably predicted even for a small number of steps, 
which further amplifies our previously disturbing conclusion.  

\section*{Discussion and conclusion}

In this paper we intended to investigate the network connectivity inference and the dynamics predictability of a high-dimensional spatiotemporal chaotic system. 
We have shown that these objectives could be achieved for a shorter prediction interval and relatively small size chaotic systems using the Tensorflow library.  
However, the investigation also unveiled an unexpected and undesirable result, showing that the tool we were using is unreliable for this task. 

Our further investigation has shown that the TensorFlow model learning part is excellent, and it provides very good inferences of the unknown network connectivity and biases. 
We suspect that the culprit in this investigation 
is the 'model.predict()' method, which seems nondeterministic. We agree and we support the fact that efficiently training neural networks requires randomness, since it is based on various stochastic versions 
of the backpropagation algorithm, however we do not understand why the 'model.predict()' method has such an 'unpredictable' behavior. We tried to fix this by setting the system environment variable 
'TF\_DETERMINISTIC' to '1', as recommended in some documentation sources, however that didn't changed anything. We also tried to use 'tf.config.experimental.enable\_op\_determinism()', but again this didn't work out either. 
Also, we should mention that in all these experiments we were only using CPU computing, so no floating point issues arising from GPUs are expected. 
Moreover, our results show that this unpredictability of the 'model.predict()' method increases very fast with the size of the model, which could have very bad effects on attempts to predict the behavior 
of important high-dimensional spatiotemporal systems arising in pretty much every scientific field: physics, biology, medicine, economics, meteorology etc. According to our results, the model inferred using 
TensorFlow could be very good (almost perfect), however it is the 'model.predict()' method that subsequently fails when it is applied iteratively, at least in the particular cases presented here. 
We would like to note here that we have obtained similar results for multilayer neural networks, showing once again that it is not the learning process in TensorFlow that fails, but the 'model.predict()' method. 
We didn't included these multilayer results here since they are quite similar, and we wanted to keep our report simple and short.

In conclusion, we should stress that many scientific problems require the prediction of chaotic systems, which by the way are deterministic and should not include sources of randomness, unless a random perturbation is required 
by the problem to be solved, and in these cases the TensorFlow 'model.predict()' method does not seem to be the appropriate method to use, since we end up predicting a chaotic system with another chaotic system.
For the record, in this investigation we have used 'tensorflow-cpu v.2.13.0', and 
hopefully in the future these annoying issues of an otherwise great tool will be resolved. 
In the end we should mention an amusing analogy of this work with a quote from the famous "The Magnificent Seven" movie:

Chico: "Ah, that was the greatest shot I've ever seen."

Britt: "The worst! I was aiming at the horse."

\section*{Appendix 1}

\begin{footnotesize}
\begin{verbatim}
import os
import numpy as np
import matplotlib.pyplot as plt
import tensorflow as tf
from tensorflow import keras
from tensorflow.keras import layers

def numpy_predict(x0,w,b,N,T):
    x = np.zeros((T,N))
    x[0] = x0
    for t in range(T-1):
        x[t+1] = np.tanh(np.dot(w,x[t]) + b)
    return x

def make_tf_model(w,b,N):
    model = keras.Sequential()
    model.add(layers.Input(shape=(N,)))
    model.add(layers.Dense(N, activation="tanh"))
    model.layers[0].set_weights([w.T,b])
    return model

def tf_model_predict(x0,model,N,T):
    x = np.zeros((T,N))
    x[0] = x0
    for t in range(T-1):
        x[t+1] = model.predict(x[t].reshape(1,N),verbose=0)
    return x

def plot_array(x,fname):
    fig = plt.figure(figsize=(10,5))
    im = plt.imshow(x.T, cmap="gray", interpolation="nearest",vmin=-1, vmax=1)
    plt.colorbar(im)
    plt.xlabel('$t$', fontsize=12)
    plt.ylabel('$x$', fontsize=12)
    fig.savefig(fname, bbox_inches='tight')

if __name__ == "__main__":
    N,T = 50,100
    np.random.seed(123); tf.random.set_seed(123)
    x0 = 2*np.random.rand(N) - 1
    b = 2*np.random.rand(N) - 1
    w = 2*np.random.rand(N,N) - 1
    x = numpy_predict(x0,w,b,N,T)
    model = make_tf_model(w,b,N)
    y = tf_model_predict(x0,model,N,T)
    err = np.linalg.norm(y-x)*100.0/np.linalg.norm(x)
    print("err=",err,"%")
    plot_array(x,"numpy.pdf") 
    plot_array(y,"tensorflow.pdf")
    plot_array(y-x,"difference.pdf")
\end{verbatim}
\end{footnotesize}

\section*{Appendix 2}

\begin{footnotesize}
\begin{verbatim}
import os
import numpy as np
import matplotlib.pyplot as plt
import tensorflow as tf
from tensorflow import keras
from tensorflow.keras import layers

def numpy_predict(x0,w,b,N,T):
    x = np.zeros((T,N))
    x[0] = x0
    for t in range(T-1):
        x[t+1] = np.tanh(np.dot(w,x[t]) + b)
    return x

def make_tf_model(w,b,N):
    model = keras.Sequential()
    model.add(layers.Input(shape=(N,)))
    model.add(layers.Dense(N, activation="tanh"))
    model.layers[0].set_weights([w.T,b])
    return model

def tf_model_predict(x0,model,N,T):
    x = np.zeros((T,N))
    x[0] = x0
    for t in range(T-1):
        x[t+1] = model.predict(x[t].reshape(1,N),verbose=0)
    return x

def plot_f(x,fname):
    t = [i for i in range(len(x))]
    fig = plt.figure(figsize=(10,5))
    plt.plot(t,x,color="black",linewidth=1.0)
    plt.xlabel('$t$', fontsize=12)
    plt.ylabel('$\epsilon$ %', fontsize=12)
    plt.xlim(0,len(x))
    plt.ylim(0,1.05*np.max(x))
    plt.text(85, 20, "$N=50$", fontsize=12)
    fig.savefig(fname, bbox_inches='tight')

if __name__ == "__main__":
    N,T,M = 50,100,100
    np.random.seed(123); tf.random.set_seed(123)
    f = np.zeros(T)
    for m in range(M):
        print("m=",m)
        x0 = 2*np.random.rand(N) - 1
        b = 2*np.random.rand(N) - 1
        w = 2*np.random.rand(N,N) - 1
        x = numpy_predict(x0,w,b,N,T)
        model = make_tf_model(w,b,N)
        y = tf_model_predict(x0,model,N,T)
        z = y - x
        for t in range(1,T):
            f[t] += np.linalg.norm(z[t])*100.0/np.linalg.norm(x[t])
    f = f/M
    plot_f(f,"fig_err.pdf")
\end{verbatim}
\end{footnotesize}


\begin{thebibliography}{99}

\bibitem{key-1} 
M. Andrecut, S. Kauffman, \textit{Phase transition in a class of nonlinear random networks}, Phys. Rev. E 82, 022105 (2010).

\bibitem{key-2} 
M. Abadi et al., \textit{TensorFlow: A system for large-scale machine learning},
12th USENIX Symposium on Operating Systems Design and Implementation (OSDI 16), USENIX Association (2016), pp. 265-283

\bibitem{key-3} 
https://www.tensorflow.org

\bibitem{key-4} 
https://keras.io
				
\end{thebibliography}
\end{document}